# Cognitive Surveillance: Why does it never appear among the AVSS Conferences topics?

Emanuel Diamant
VIDIA-mant, POB 933, Kiriat Ono 5510801, Israel
emanl.245@gmail.com

**Abstract:** Video Surveillance is a fast evolving field of research and development (R&D) driven by the urgent need for public security and safety (due to the growing threats of terrorism, vandalism, and anti-social behavior). Traditionally, surveillance systems are comprised of two components – video cameras distributed over the guarded area and human observer watching and analyzing the incoming video. Explosive growth of installed cameras and limited human operator's ability to process the delivered video content raise an urgent demand for developing surveillance systems with human like cognitive capabilities, that is – Cognitive surveillance systems. The growing interest in this issue is testified by the tens of workshops, symposiums and conferences held over the world each year. The IEEE International Conference on Advanced Video and Signal-Based Surveillance (AVSS) is certainly one of them. However, for unknown reasons, the term Cognitive Surveillance does never appear among its topics. As to me, the explanation for this is simple – the complexity and the indefinable nature of the term "Cognition". In this paper, I am trying to resolve the problem providing a novel definition of cognition equally suitable for biological as well as technological applications. I hope my humble efforts will be helpful.

## 1. Introduction

Over the last 10-15 years or so, video surveillance has become a very hot topic in public, private and homeland security issues. The driving force behind this rise of interest in video surveillance is the growing public concern about the threats posed to the modern society by terrorism, crime, suspicious activities and vandalism. This growing demand for security is justified by the memory of the recent and past events: on September 11, 2001 in New York, on March 11, 2004 in Madrid, on July 7, 2005 in London and the latest event during the Boston Marathon on April 15, 2013.

Developing intelligent and robust surveillance systems has become thus the main focus of interest and a challenge for the research and engineering communities. Being supported by ample funding, which comes both from governments and industries, the main R&D efforts are aimed at creating technologically advanced surveillance systems, which will enable to deter the anticipated threats and provide the required levels of security and safety.

But viewing video surveillance as only a technological challenge may be a slip up. After all, visual surveillance is a behavioral function that has been with us for hundreds of millions of years. In course of natural evolution, the life of every living being has always revolved around two main issues: to look for food to eat and to look for how to avoid to be eaten by something else. Both tasks imply a constant interaction with the environment – a constant surveillance of the surrounding area. It is compulsory for every single living being, but for small, group-living animals which have to search for their food on the ground, the presence of one or more members of the group which, instead of feeding, remain vigilant for predators was certainly of great advantage since it had allowed the remainder of the group to forage undisturbed and therefore more efficiently [3]. Examples of such behavioral practice have been widely reported in biological literature

The most often studied surveillance strategies (in biology they call them vigilant, sentinel, "watchman" and so on behavioral strategies) are described for: dwarf mongooses colonies [1], bird flocks [2], meerkats (Suricata suricatta) [3] and various kinds of monkeys. Meerkats are an especially interesting case: Meerkats are obligatory cooperative animals with labor division and specific role allocation habits. The members of the group specialize in one of the roles, such as babysitting the young, foraging, food sharing and guard duty. During the day, each member of the group takes a turn at guard duty by pausing briefly to stand erect and scan the skies for predators. His constant peeping lets the group know that the skies are clear. That resembles role sharing and specialization in human societies when specially trained and equipped (with television cameras) society members are watching their surroundings thus performing the required surveillance duties.

This short digression into the biological origins of surveillance is made with only one purpose in mind – to underscore that surveillance is a mental (brain) function that facilitates animal and human threat avoidance and survival. It is commonly agreed that surveillance is enabled by a couple of other brain functions such as perception, reasoning, decision making and action taking, collectively designated as cognitive functions. From here, a transition toward a notion of "Cognitive Surveillance" is self-evident.

The idea to use a biological function developed by living creatures (in course of their natural evolution), as a blueprint for technological problem solving is definitely not new and has been widely accepted in scientific and engineering communities. Therefore, Video Surveillance R&D is not an exception here. Under the nickname "Bio-inspired Cognitive Surveillance" the subject has become the pivotal theme of the research underpinning the Fourth Generation surveillance systems design efforts [4]. A vast number of publications devoted to the subject can be mentioned now, and the items [5], [6], [7] in the reference list are given only as an example, which do not exhaust the real volume of relevant publications.

However, despite the growing amount of research in the field, the bio-inspired approach does not take off and does not yield the expected results yet. The reason for this is simple – bio-inspired approaches in general, and cognitive surveillance in particular, are usually preoccupied with (and try to imitate) the particular biological sub-functions (relevant to a specific case), rather than to investigate and to try to comprehend the complex general function that is looming behind the partial constituents. In the case of cognitive surveillance, the research efforts are directed to the investigation of various cognitive "mechanisms", such as perception, reasoning, memory, learning, and so on, while any attempt to understand what cognition is in general and what does it mean to be cognitive – any such attempt is being permanently discarded.
I don't think that that is a wise way to deal with cognitive surveillance. I would like to do this in another manner.

## 2. So, what cognition is and what it does?

To begin this part of the discourse, I would like to provide a quote from Pamela Lion's paper "The biogenic approach to cognition": "After half a century of cognitive revolution we remain far from agreement about what cognition is and what cognition does. It was once thought that these questions could wait until the data were in. Today there is a mountain of data, but no way of making sense of it." [8].

Despite of this, or, may be, just because of this, the adjective "cognitive" has become extremely popular in our lexicon and it is widely used in a huge amount of denominations: Cognitive science, Cognitive biology, Cognitive behavior, Cognitive neuroscience – to recall only some of the popular labels on the biological part of the list. And – Cognitive information, Cognitive vision, Cognitive robotics, Cognitive computing and, of course, Cognitive surveillance – the well known brand names populating the technical part of the list. Google's inquiry for every one of the mentioned above items returns the following numbers of acknowledgement: Cognitive science - about 1,610,000 results, Cognitive neuroscience - about 796,000 results, Cognitive behavior - about 407,000 results, Cognitive biology - about 26,400 results. That is on the one hand; on the other hand: Cognitive computing - about 331,000 results, Cognitive information - about 160,000 results, Cognitive robotics - about 84,800 results, Cognitive vision - about 44,500 results, and Cognitive surveillance - about 3,030 results.

It is not surprising that Cognitive surveillance is mentioned only in 3,030 cases while Cognitive computing comes out in about 331,000 cases. After all, Cognitive surveillance is an emerging concept and Cognitive computing is the most visible sign of the forthcoming technological revolution. Forecasting the future, IBM has nominated Cognitive computing as the new generation of computers that has the potential to change the world [9]. Shweta Dubey, in agreement with many other research analysts, writes in her comment ("Is Cognitive Computing the Next Disruptive Technology?"): "Cognitive computing will be a larger part of the future as an emerging field in which computers can operate more like a human brain" [10].

The expectation for Cognitive computing marvels stems from an urgent need to handle massive amounts of data which are being generated and stored ubiquitously in every discipline and every aspect of our life. This data deluge is known today as the Big Data era. It is generally believed that Cognitive computing will be the right way to meet the challenges posed by accelerated growth of the Big Data volumes.

In this regard, it must be mentioned that video surveillance data is a very significant part of the Big Data heap.
According to the International Data Corporation's recent report, "The Digital Universe in 2020," half of global big data was surveillance video in 2012, and the percentage is set to increase to 65 percent by 2015 [11].

Therefore, it is right to assume that the slow inauguration of cognitive surveillance can be explained as follows – cognitive surveillance R&D is just waiting when advances in cognitive computing will pave for them the way for human like (cognitive) video data handling and processing!

However, among biological scientists the term "cognitive" has nothing to do with "data" and is definitely associated with "information" (handling and processing). The latest (April 2013) paper articulates this generally accepted point of view in the following way: "Nervous systems are standardly interpreted as information processing input–output devices" [12]. Wikipedia also does not let any place for doubts: "…cognition is typically assumed to be information processing in a participant's or operator's mind or brain. Cognition is a faculty for the processing of information, applying knowledge, and changing preferences" [13].

Differences in opinion about data-information dichotomy are not new (please see, for example, the arguments in Floridi's paper "Is Semantic Information Meaningful Data?" [14]). I do not intend to take part in this rivalry. As far as I understand the problem, the right thing to do will be to try and to dissect this dichotomy by providing a serious investigation into "What information really is?"

## 3. What is information?

It is generally agreed that a consensus definition of information does not exist. Therefore, I would like to propose a definition of my own (borrowed and extended from the Kolmogorov's definition of information first introduced in the mid-sixties of the past century): **Information is a linguistic description of structures observable in a given data set** [15].

Two types of structures could be distinguished in a data set – primary and secondary data structures. The first are data element aggregations (data clusters) whose agglomeration is guided by similarity in some physical properties; the others are compositions of primary data structures (that is, secondary data structures), which appear in the observer's brain. The arrangement and grouping of secondary data structures is guided by the observer's customs and habits. Therefore, the primary data structures could be called Physical data structures and the secondary data structures – Meaningful or Semantic data structures. And their descriptions should be called accordingly **Physical Information** and **Semantic Information**.

This subdivision is usually overlooked in the contemporary information/data processing approaches leading to mistaken and erroneous data handling methods and techniques.

In the sake of paper space saving, I would advise the interested readers to go to my website (http://www.vidia-mant.info ), where a full list of my recent relevant publications on the subject is available and a more extended explanation of information description duality can be found. Meanwhile I will try to explain the consequences that immediately pop up from this new understanding. Relying on the just acquired notion of information, we can move further to a more suitable definition of cognition. In the light of our new knowledge, we can certainly posit that **cognition is the ability to process information**. And that is what our brains are doing – the brains are processing information, not data (as it is usually assumed). And that is what we are striving to replicate in our Cognitive Systems designs – the sacred information processing.

In the light of the new knowledge, we can now definitely say that physical information is carried by the data and therefore can be promptly extracted from it. At the same time, semantic information is a description of observer's arrangement of physical data structures and therefore has nothing to do with data, because semantics is not a property of the data, it is a property of an observer that is watching and scrutinizing the data. As such, semantics is always subjective and it is always a result of mutual agreements and conventions that are established in a certain group of observers, or a future companionship of a cognitive machine and a human, which are performing as a team and share a common understanding (a common semantic information) about their environment. An important sequel of this is that the semantic information can not be learned autonomously, but it should be provided to a cognitive system from the outside – semantics has to be taught and not learned (as it is usually assumed by cognitive system designers). In this regard, it must be mentioned that the emerging practice to interchange the semantic information hierarchy with a learned ontology is a wrong way to do the right thing of providing a cognitive system with the needed reference knowledge.

Another important corollary that follows from the new understanding of information is that an information description is always a linguistic description, that is, it is a string of symbols which can take a form of a mathematical formula (don't forget that mathematics is a sort of a language) or a natural language item – a word, a sentence, or a piece of text. That is a very important outcome of the new theory. Considering that contemporary approaches to the problem of information processing are assuming computer involvement in the processing task, it must be certainly clarified: contemporary computers are data processing machines which are not suitable for processing natural language texts (which are semantic information carriers).

(Inaccurate use of adjective "cognitive" in various new compound words often leads to unexpected and amusing results. For example, "Cognitive computing": Here "cognitive" implies information processing while "computing" implies data processing; in conjunction – a classical oxymoron, something like "exact estimate". At the same time, "vision" and "surveillance" are by definition cognitive procedures; so, "Cognitive vision" and "Cognitive surveillance" are widely used tautologies. But, who cares?)

Conference paper format does not allow a full-blown review of the consequences following from the new "information" and "cognitive" notions understanding. Not all of the novelties are elaborated to the needed extent by the author and so the theme is waiting for further exploration by the research community.

## 4. Conclusions

Success in Cognitive Surveillance system design can be reached only on the ground of new understanding about what is information and, consequently, what is information processing and cognition. Otherwise designers' efforts are doomed to failure. Unfortunately, this is not the state of affairs today and even the term Cognitive Surveillance does not appear among the topics of many prestigious gatherings and conferences. The 11-th AVSS Conference is not an exception here. However, Scholar Google inquiry for Cognitive Surveillance returns 129 results – and that is a sign that the Cognitive Surveillance theme is not entirely discarded from public discourse and that a hope is still there for a creative and fruitful future research accomplishments.

## 5. References

[1] Anne E. Rasa, **Coordinated Vigilance in Dwarf Mongoose Family Groups: The 'Watchman's Song' Hypothesis and the Costs of Guarding**, Ethology, Volume 71, Issue 4, pages 340–344, January-December 1986

[2] Linda I. Holle´n, Matthew B.V. Bell, and Andrew N. Radford, **Cooperative Sentinel Calling? Foragers Gain Increased Biomass Intake**, Current Biology 18, 576–579, April 22, 2008, Available: http://ac.els-

cdn.com/S0960982208003151/1-s2.0-S0960982208003151-main.pdf?_tid=442b34a0-8c9f-11e3-9a49-00000aab0f02&acdnat=1391410322_af43a2bae894e1b3dabe0f826fa4eb8b

[3] Manser, M.B., **Response of foraging group members to sentinel calls in suricates, Suricata suricatta**. Proc. R. Soc. Lond. B. Biol. Sci. 266, 1999, pages 1013–1019.
http://www.ncbi.nlm.nih.gov/pmc/articles/PMC1689937/pdf/9WPJG7VVH0E9LPFW_266_1013.pdf

[4] Matteo Pinasco and Carlo S. Regazzoni, **Video surveillance and bio-inspired embodied cognitive systems,** First Workshop on Video Surveillance projects in Italy 2008, pp. 36-38.
http://imagelab.ing.unimore.it/visit2008/AttiVISIT2008.pdf

[5] D. Vernon, G. Metta and G. Sandini, **A survey of artificial cognitive systems: Implications for the autonomous development of mental capabilities in computational agents**, IEEE Transactions on Evolutionary Computation 11(2) (2007), 151–180,
http://pasa.cognitivehumanoids.eu/pasapdf/262_Vernon_etal2007.pdf

[6] Alessio Dore, Matteo Pinasco, Lorenzo Ciardelli and Carlo Regazzoni, **A bio-inspired system model for interactive surveillance applications,** Journal of Ambient Intelligence and Smart Environments 0 (2011) 1–17 1, IOS Press. http://www.isip40.it/resources/papers/2011/AISE11.pdf

[7] Regazzoni, C.S., **Bio-inspired autonomous dynamical systems for interactive and cognitive environments,** Computer Engineering & Systems (ICCES), 2012 Seventh International Conference on
http://ieeexplore.ieee.org/xpls/abs_all.jsp?arnumber=6408461

[8] Pamela Lyon, **The biogenic approach to cognition,** Cogn Process (2005)
http://hss.adelaide.edu.au/philosophy/cogbio/publications/Lyon_2006_biogenic_penult.pdf

[9] Brian Deagon, **IBM Predicts Cognitive Systems As New Computing Wave**, Investor's Business Daily,
http://news.investors.com/technology-tech-exec-qanda/012313-641600-ibm-cognitive-computers-future-trend-5-senses.htm?p=full

[10] Shweta Dubey, **Is Cognitive Computing the Next Disruptive Technology?** January 7, 2013
http://beta.fool.com/shwetadubey/2013/01/07/cognitive-computing-next-disruptive-technology/20111/

[11] Tiejun Huang,  **Surveillance Video: The Biggest Big Data,** Computer, February 2014.
http://www.computer.org/portal/web/computingnow/archive/february2014?lf1=859719755f993516066397d16868842

[12] Fred Keijzer, Marc van Duijn, Pamela Lyon, **What nervous systems do: early evolution, input–output, and the skin brain thesis,** Adaptive Behavior, April 2013, vol. 21, no. 2, pp. 67-85
http://adb.sagepub.com/content/21/2/67

[13] **Cognition**, From Wikipedia, the free encyclopedia, http://en.wikipedia.org/wiki/Cognition

[14] Luciano Floridi, **Is Semantic Information Meaningful Data?,** Philosophy and Phenomenological Research, Vol. LXX, No. 2, March 2005, http://philsci-archive.pitt.edu/2536/1/iimd.pdf

[15} Diamant, E., **Let Us First Agree on what the Term "Semantics" Means: An Unorthodox Approach to an Age-Old Debate**, In M. T. Afzal, Ed., "Semantics - Advances in Theories and Mathematical Models",(pp. 3 – 16), InTech Publisher, 2012. Available:
http://www.intechopen.com/statistics/35998